\documentclass[conference]{IEEEtran}
\ifCLASSINFOpdf
\else
\fi
\usepackage{xcolor}
\definecolor{mygreen}{rgb}{0.13, 0.55, 0.13}
\usepackage{booktabs}
\usepackage{graphicx}
\usepackage{algorithm2e,algorithmic}
\usepackage{multirow}
\usepackage{amsfonts}
\usepackage{subfigure}

\newcommand{\spotwo}{SpO$_2$\;}

\newcommand{\ie}{i.e.,\;}
\newcommand{\eg}{e.g.,\;}

\newtheorem{definition}{Definition}

\begin{document}
%
\title{Curating Naturally Adversarial Datasets for Learning-Enabled Medical Cyber-Physical Systems}

\author{
\IEEEauthorblockN{Sydney Pugh}
\IEEEauthorblockA{
University of Pennsylvania\\
Philadelphia, PA 19104\\
sfpugh@seas.upenn.edu}
\and
\IEEEauthorblockN{Ivan Ruchkin}
\IEEEauthorblockA{
University of Florida\\
Gainesville, FL 32611\\
iruchkin@ece.ufl.edu}
\and
\IEEEauthorblockN{James Weimer}
\IEEEauthorblockA{
Vanderbilt University\\
Nashville, TN 37235\\
james.weimer@vanderbilt.edu}
\and
\IEEEauthorblockN{Insup Lee}
\IEEEauthorblockA{
University of Pennsylvania\\
Philadelphia, PA 19104\\
lee@seas.upenn.edu}}


%


\maketitle

\begin{abstract}
In medical cyber-physical systems (CPS), where patient safety is a top priority, the robustness of learning-enabled components (LECs) becomes crucial.
Therefore, a comprehensive robustness evaluation is necessary for the successful deployment of these systems.
Existing research predominantly focuses on robustness to \emph{synthetic adversarial examples}, crafted by adding imperceptible perturbations to clean input data. However, these synthetic adversarial examples do not accurately reflect the most challenging real-world scenarios, especially in the context of healthcare data. Consequently, robustness to synthetic adversarial examples may not necessarily translate to robustness against \emph{naturally occurring adversarial examples}. 
We propose a method to curate datasets comprised of natural adversarial examples to evaluate the robustness of LECs. The method relies on probabilistic labels obtained from automated weakly-supervised labeling that combines noisy and cheap-to-obtain labeling heuristics. Based on these labels, our method adversarially orders the input data and uses this ordering to construct a sequence of increasingly adversarial datasets. Our evaluation on six medical CPS case studies and three non-medical case studies demonstrates the efficacy and statistical validity of our approach to generating naturally adversarial datasets.
\end{abstract}


%
\IEEEpeerreviewmaketitle

\def\ah{\hat{a}}
\def\bh{\hat{b}}
\def\ch{\hat{c}}
\def\dh{\hat{d}}
\def\eh{\hat{e}}
\def\fh{\hat{f}}
\def\gh{\hat{g}}
\def\hh{\hat{h}}
\def\ih{\hat{i}}
\def\jh{\hat{j}}
\def\kh{\hat{k}}
\def\lh{\hat{l}}
\def\mh{\hat{m}}
\def\nh{\hat{n}}
\def\oh{\hat{o}}
\def\ph{\hat{p}}
\def\qh{\hat{q}}
\def\rh{\hat{r}}
\def\sh{\hat{s}}
\def\th{\hat{t}}
\def\uh{\hat{u}}
\def\vh{\hat{v}}
\def\wh{\hat{w}}
\def\xh{\hat{x}}
\def\yh{\hat{y}}
\def\zh{\hat{z}}

\def\ab{\bar{a}}
\def\bb{\bar{b}}
\def\cb{\bar{c}}
\def\db{\bar{d}}
\def\eb{\bar{e}}
\def\fb{\bar{f}}
\def\gb{\bar{g}}
\def\hb{\bar{h}}
\def\ib{\bar{i}}
\def\jb{\bar{j}}
\def\kb{\bar{k}}
\def\lb{\bar{l}}
\def\mb{\bar{m}}
\def\nb{\bar{n}}
\def\ob{\bar{o}}
\def\pb{\bar{p}}
\def\qb{\bar{q}}
\def\rb{\bar{r}}
\def\sb{\bar{s}}
\def\tb{\bar{t}}
\def\ub{\bar{u}}
\def\vb{\bar{v}}
\def\wb{\bar{w}}
\def\xb{\bar{x}}
\def\yb{\bar{y}}
\def\zb{\bar{z}}

\def\at{\tilde{a}}
\def\bt{\tilde{b}}
\def\ct{\tilde{c}}
\def\dt{\tilde{d}}
\def\et{\tilde{e}}
\def\ft{\tilde{f}}
\def\gt{\tilde{g}}
\def\htt{\tilde{h}}
\def\it{\tilde{i}}
\def\jt{\tilde{j}}
\def\kt{\tilde{k}}
\def\lt{\tilde{l}}
\def\mt{\tilde{m}}
\def\nt{\tilde{n}}
\def\ot{\tilde{o}}
\def\pt{\tilde{p}}
\def\qt{\tilde{q}}
\def\rt{\tilde{r}}
\def\st{\tilde{s}}
\def\tt{\tilde{t}}
\def\ut{\tilde{u}}
\def\vt{\tilde{v}}
\def\wt{\tilde{w}}
\def\xt{\tilde{x}}
\def\yt{\tilde{y}}
\def\zt{\tilde{z}}

\def\Ah{\hat{A}}
\def\Bh{\hat{B}}
\def\Ch{\hat{C}}
\def\Dh{\hat{D}}
\def\Eh{\hat{E}}
\def\Fh{\hat{F}}
\def\Gh{\hat{G}}
\def\Hh{\hat{H}}
\def\Ih{\hat{I}}
\def\Jh{\hat{J}}
\def\Kh{\hat{K}}
\def\Lh{\hat{L}}
\def\Mh{\hat{M}}
\def\Nh{\hat{N}}
\def\Oh{\hat{O}}
\def\Ph{\hat{P}}
\def\Qh{\hat{Q}}
\def\Rh{\hat{R}}
\def\Sh{\hat{S}}
\def\Th{\hat{T}}
\def\Uh{\hat{U}}
\def\Vh{\hat{V}}
\def\Wh{\hat{W}}
\def\Xh{\hat{X}}
\def\Yh{\hat{Y}}
\def\Zh{\hat{Z}}

\def\Ab{\bar{A}}
\def\Bb{\bar{B}}
\def\Cb{\bar{C}}
\def\Db{\bar{D}}
\def\Eb{\bar{E}}
\def\Fb{\bar{F}}
\def\Gb{\bar{G}}
\def\Hb{\bar{H}}
\def\Ib{\bar{I}}
\def\Jb{\bar{J}}
\def\Kb{\bar{K}}
\def\Lb{\bar{L}}
\def\Mb{\bar{M}}
\def\Nb{\bar{N}}
\def\Ob{\bar{O}}
\def\Pb{\bar{P}}
\def\Qb{\bar{Q}}
\def\Rb{\bar{R}}
\def\Sb{\bar{S}}
\def\Tb{\bar{T}}
\def\Ub{\bar{U}}
\def\Vb{\bar{V}}
\def\Wb{\bar{W}}
\def\Xb{\bar{X}}
\def\Yb{\bar{Y}}
\def\Zb{\bar{Z}}

\def\At{\tilde{A}}
\def\Bt{\tilde{B}}
\def\Ct{\tilde{C}}
\def\Dt{\tilde{D}}
\def\Et{\tilde{E}}
\def\Ft{\tilde{F}}
\def\Gt{\tilde{G}}
\def\Ht{\tilde{H}}
\def\It{\tilde{I}}
\def\Jt{\tilde{J}}
\def\Kt{\tilde{K}}
\def\Lt{\tilde{L}}
\def\Mt{\tilde{M}}
\def\Nt{\tilde{N}}
\def\Ot{\tilde{O}}
\def\Pt{\tilde{P}}
\def\Qt{\tilde{Q}}
\def\Rt{\tilde{R}}
\def\St{\tilde{S}}
\def\Tt{\tilde{T}}
\def\Ut{\tilde{U}}
\def\Vt{\tilde{V}}
\def\Wt{\tilde{W}}
\def\Xt{\tilde{X}}
\def\Yt{\tilde{Y}}
\def\Zt{\tilde{Z}}

\def\alphah{\hat{\alpha}}
\def\alphab{\bar{\alpha}}
\def\alphat{\tilde{\alpha}}

\def\betah{\hat{\beta}}
\def\betab{\bar{\beta}}
\def\betat{\tilde{\beta}}

\def\gammah{\hat{\gamma}}
\def\gammab{\bar{\gamma}}
\def\gammat{\tilde{\gamma}}

\def\deltah{\hat{\delta}}
\def\deltab{\bar{\delta}}
\def\deltat{\tilde{\delta}}
\def\Deltah{\hat{\Delta}}
\def\Deltab{\bar{\Delta}}
\def\Deltat{\tilde{\Delta}}

\def\etah{\hat{\eta}}
\def\etab{\bar{\eta}}
\def\etat{\tilde{\eta}}

\def\thetah{\hat{\theta}}
\def\thetab{\bar{\theta}}
\def\thetat{\tilde{\theta}}

\def\lambdah{\hat{\lambda}}
\def\lambdab{\bar{\lambda}}
\def\lambdat{\tilde{\lambda}}
\def\Lambdah{\hat{\Lambda}}
\def\Lambdab{\bar{\Lambda}}
\def\Lambdat{\tilde{\Lambda}}

\def\muh{\hat{\mu}}
\def\mub{\bar{\mu}}
\def\mub{\tilde{\mu}}

\def\pih{\hat{\pi}}
\def\pib{\bar{\pi}}
\def\pit{\tilde{\pi}}

\def\rhoh{\hat{\rho}}
\def\rhob{\bar{\rho}}
\def\rhot{\tilde{\rho}}

\def\sigmah{\hat{\lambda}}
\def\sigmab{\bar{\sigma}}
\def\sigmat{\tilde{\sigma}}
\def\Sigmab{\bar{\Sigma}}
\def\Sigmah{\hat{\Sigma}}
\def\Sigmat{\tilde{\Sigma}}

\def\phih{\hat{\phi}}
\def\phib{\bar{\phi}}
\def\phit{\tilde{\phi}}

\def\psih{\hat{\psi}}
\def\psib{\bar{\psi}}
\def\psit{\tilde{\psi}}

\def\omegah{\hat{\omega}}
\def\omegab{\bar{\omega}}
\def\omegat{\tilde{\omega}}
\def\Omegah{\hat{\Omega}}
\def\Omegab{\bar{\Omega}}
\def\Omegat{\tilde{\Omega}}

\def\As{\mathcal{A}}
\def\Bs{\mathcal{B}}
\def\Cs{\mathcal{C}}
\def\Ds{\mathcal{D}}
\def\Es{\mathcal{E}}
\def\Fs{\mathcal{F}}
\def\Gs{\mathcal{G}}
\def\Hs{\mathcal{H}}
\def\Is{\mathcal{I}}
\def\Js{\mathcal{J}}
\def\Ks{\mathcal{K}}
\def\Ls{\mathcal{L}}
\def\Ms{\mathcal{M}}
\def\Ns{\mathcal{N}}
\def\Os{\mathcal{O}}
\def\Ps{\mathcal{P}}
\def\Qs{\mathcal{Q}}
\def\Rs{\mathcal{R}}
\def\Ss{\mathcal{S}}
\def\Ts{\mathcal{T}}
\def\Us{\mathcal{U}}
\def\Vs{\mathcal{V}}
\def\Ws{\mathcal{W}}
\def\Xs{\mathcal{X}}
\def\Ys{\mathcal{Y}}
\def\Zs{\mathcal{Z}}

\def\Ash{\hat{\mathcal{A}}}
\def\Bsh{\hat{\mathcal{B}}}
\def\Csh{\hat{\mathcal{C}}}
\def\Dsh{\hat{\mathcal{D}}}
\def\Esh{\hat{\mathcal{E}}}
\def\Fsh{\hat{\mathcal{F}}}
\def\Gsh{\hat{\mathcal{G}}}
\def\Hsh{\hat{\mathcal{H}}}
\def\Ish{\hat{\mathcal{I}}}
\def\Jsh{\hat{\mathcal{J}}}
\def\Ksh{\hat{\mathcal{K}}}
\def\Lsh{\hat{\mathcal{L}}}
\def\Msh{\hat{\mathcal{M}}}
\def\Nsh{\hat{\mathcal{N}}}
\def\Osh{\hat{\mathcal{O}}}
\def\Psh{\hat{\mathcal{P}}}
\def\Qsh{\hat{\mathcal{Q}}}
\def\Rsh{\hat{\mathcal{R}}}
\def\Ssh{\hat{\mathcal{S}}}
\def\Tsh{\hat{\mathcal{T}}}
\def\Ush{\hat{\mathcal{U}}}
\def\Vsh{\hat{\mathcal{V}}}
\def\Wsh{\hat{\mathcal{W}}}
\def\Xsh{\hat{\mathcal{X}}}
\def\Ysh{\hat{\mathcal{Y}}}
\def\Zsh{\hat{\mathcal{Z}}}

\def\Ast{\tilde{\mathcal{A}}}
\def\Bst{\tilde{\mathcal{B}}}
\def\Cst{\tilde{\mathcal{C}}}
\def\Dst{\tilde{\mathcal{D}}}
\def\Est{\tilde{\mathcal{E}}}
\def\Fst{\tilde{\mathcal{F}}}
\def\Gst{\tilde{\mathcal{G}}}
\def\Hst{\tilde{\mathcal{H}}}
\def\Ist{\tilde{\mathcal{I}}}
\def\Jst{\tilde{\mathcal{J}}}
\def\Kst{\tilde{\mathcal{K}}}
\def\Lst{\tilde{\mathcal{L}}}
\def\Mst{\tilde{\mathcal{M}}}
\def\Nst{\tilde{\mathcal{N}}}
\def\Ost{\tilde{\mathcal{O}}}
\def\Pst{\tilde{\mathcal{P}}}
\def\Qst{\tilde{\mathcal{Q}}}
\def\Rst{\tilde{\mathcal{R}}}
\def\Sst{\tilde{\mathcal{S}}}
\def\Tst{\tilde{\mathcal{T}}}
\def\Ust{\tilde{\mathcal{U}}}
\def\Vst{\tilde{\mathcal{V}}}
\def\Wst{\tilde{\mathcal{W}}}
\def\Xst{\tilde{\mathcal{X}}}
\def\Yst{\tilde{\mathcal{Y}}}
\def\Zst{\tilde{\mathcal{Z}}}

\def\E{\mathbf{E}}
\let\Para\P
\def\P{\mathbf{P}}
\def\Cov{\textbf{Cov}}
\def\Cor{\textbf{Cor}}
\def\acc{\textsc{acc}}
\def\pearson{\textsc{Pearson}}

\newcommand{\ind}[1]{\mathbf{1}\left(#1\right)}
\section{Introduction}
\label{sec:intro}

\begin{figure*}[t]
    \centering
    \includegraphics[width=\linewidth]{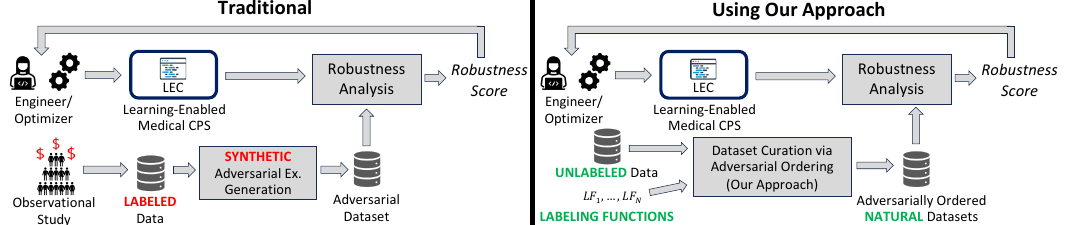}
    \caption{Traditional robustness evaluation versus our approach. LEC = Learning-Enabled Component.}
    \label{fig:mcps-loop}
\end{figure*}


Deep learning models have demonstrated remarkable performance in the analysis of medical time-series data~\cite{faust2018deep,morid2023timeseries}, inciting substantial interest in the development of learning-enabled medical cyber-physical systems (CPS)~\cite{lee2010medical,watson2022medical}. 
For example, physiologic monitoring systems are an essential CPS for healthcare that continuously measure patient vitals and raise alarms when the vitals appear abnormal. Unfortunately, such systems are known to overwhelm caregivers with many in-actionable or non-informative alarms~\cite{paine2016systematic,drew2014insights}. 
Recent research have proposed novel learning-based algorithms for monitoring and suppressing unnecessary alarms~\cite{lameski2017suppression,nguyen2018reducing,au2019reduction}. Thus a plausible next step would be to integrate such learning-based alarm suppression algorithms into a physiologic monitoring system.
However, patient safety is a primary concern for medical CPS and thus the learning-enabled components must function properly on expected and unexpected inputs~\cite{gatouillat2018internet}. 
For instance, a learning-enabled physiologic monitoring system should be able to handle patients with abnormal physiology due to illness or instances where the sensor data is noisy due to artifact or missing due to sensor disconnections.
Successful deployment of learning-enabled medical CPS, therefore, is contingent on a thorough evaluation of the system's robustness. 

Robustness is typically evaluated by observing a model's performance against adversarial examples, particularly \emph{synthetic} adversarial examples, that are generated by intentionally adding small perturbations to \emph{labeled} input data to cause misclassification while being nearly imperceptible to humans. This evaluation approach can be challenging to apply to a time-series medical CPS for two primary reasons. Firstly, synthetic adversarial examples generally do not resemble adversarial examples that would be encountered in the real world~\cite{koh2021wilds} -- simply adding random perturbations to medical data usually yields invalid and unrealistic examples. 
%
Secondly, it is time-consuming and expensive to construct highly accurate labeled datasets for generating realistic synthetic adversarial examples. A common way to do so is to collect data via an observational study~\cite{bonafide2017video} and then have domain experts manually label the data. However, such a study is a major commitment when it comes to an initial deployment of a medical CPS, in part due to the significant effort of manually labeling examples. 

\looseness=-1
Instead, we focus on \emph{natural} adversarial examples --- real patient examples that are difficult to classify~\cite{hendrycks2021natural}. 
Natural adversarial examples capture the inherent variations and uncertainties present in real-world medical data that effectively deceive models. Therefore, evaluating the robustness of medical deep learning models against such examples becomes crucial to ensuring model robustness in the real world.  We aim to identify natural adversarial examples in \emph{unlabeled} medical time-series data so that we can curate naturally adversarial \emph{datasets}.

In an ideal world, in medicical CPS, a model's accuracy would be evaluated on a labeled natural adversarial dataset -- in reality, developing labeled medical datasets is an expensive and time-consuming task. However, obtaining weak labels for historical datasets is a cheap and quick alternative~\cite{ratner2016data},
but, assessing accuracy/robustness with respect to weak labels becomes challenging due to the uncertainty surrounding the correctness of the weak labels. In our experience, weak labels for adversarial examples are prone to significant inaccuracies due to the inherent difficulty in classifying these examples. 
 Hence, robust models tend to disagree with the weak labels of natural adversarial examples because of weak labeling inaccuracies -- not model inaccuracies. Therefore, we aim to curate \emph{adversarially ordered} datasets, wherein the proportion and severity of adversarial examples gradually increase per dataset, correspondingly leading to a diminishing accuracy of each dataset's weak labels. In practice, a robust model would be expected to show decreasing accuracy with respect to our weak labels on adversarially ordered datasets.

This paper introduces a weakly-supervised method for curating adversarially ordered datasets containing natural adversarial examples. Our method can support early-stage low-cost robustness evaluations of medical CPS as depicted in Figure~\ref{fig:mcps-loop}. Note that our evaluation method is complimentary to the traditional method as it can be applied to historical data before incurring the cost of an observational study. 
A key step of our method is to probabilistically label data using \emph{weakly-supervised data labeling}. This process leverages \emph{labeling functions}, which assign labels to subsets of the data. Labeling functions are noisy: their labels can be incorrect, incomplete, or contradictory. Furthermore, labeling functions are assumed to be conditionally independent. We start with a dataset of unlabeled time series and a set of labeling functions. We then select a subset of \emph{independent labeling functions} to be used in our method via a heuristic. The outputs of the selected labeling functions are combined to yield a probabilistic label, \ie a label and corresponding confidence in it, for each sample in the dataset.

Adversarial examples are by definition difficult to classify, hence the \emph{uncertainty in the label} of an example can indicate how adversarial it is. We identify adversarial examples as those with high label uncertainty. As indication of label uncertainty, we will use label confidence scores. Consequently, they will serve as the basis for adversarial ordering.

Unfortunately, label confidences do not reflect the true accuracies of the labels: weakly-supervised data labeling techniques are generally \emph{poorly calibrated}. To overcome this issue, we construct \emph{confidence intervals} for the probabilistic labels output by these techniques to quantify the uncertainty of the labeling process. An important consideration not previously considered is the \emph{number of labeling functions} combined to produce the label. Informed by this circumstance, our intervals indicate, for a given level of confidence, the range of possible confidence values in the assigned label. Finally, using the lower bounds of confidence intervals, we order the examples by their natural adversarialness.

We validate our method on six medical case studies and three non-medical case studies. The medical case studies include five clinical alarm datasets derived from a 551-hour alarm labeling effort from Children's Hospital of Philadelphia (CHOP), and one medical imaging dataset extracted from the Open-i dataset (https://openi.nlm.nih.gov/faq) maintained by the National Institutes of Health (NIH). The three non-medical case studies include weather-related tweets, book reviews, and synthetic YouTube comments. Our evaluation demonstrates that our approach successfully generates natural adversarial datasets with statistically valid adversarial ordering.

In summary, this paper makes the following contributions:
\begin{itemize}
    \item An approach to building datasets with increasing amounts of natural adversarial examples based on confidence intervals,
    \item A method of selecting independent labeling functions to use for weakly-supervised data labeling,
    \item An evaluation of the statistical validity of our curated adversarially ordered natural datasets on nine case studies.
\end{itemize}
\section{Related Work}
\label{sec:related}

In this section we overview related work for adversarial examples and weakly-supervised data labeling in the context of medical CPS.


\subsection{Adversarial Examples for Medical CPS} 
\looseness=-1
\cite{szegedy2013intriguing} discovered the existence of \emph{adversarial examples}: inputs intentionally designed to cause machine learning models to predict incorrectly. Subsequent research has predominately focused on \emph{synthetic} adversarial examples,
which are artificial inputs specifically generated to deceive models. A common approach for generating synthetic adversarial examples is to apply adversarial perturbations to clean inputs. For example, $\ell_p$  adversarial examples are generated by perturbing an input by some worst-case distortion that is small in the $\ell_p$ sense~\cite{vorobeychik_adversarial_2018}.
The small distortion is nearly imperceptible to humans, making it challenging to detect by inspection --- but can have a significant negative impact on the behavior of a model. Unfortunately, $\ell_p$ adversarial examples are not suitable for time series medical data because adding such noise to such data typically yields invalid and unrealistic examples. Adversarial spatial transformations (\eg rotations, translations) may also be used to perturb inputs to generate synthetic adversarial examples~\cite{xiao2018spatially}. However, this approach is only applicable for data with spatial structure (like image data) and hence is not applicable for time-series data common in medical CPS.

An alternative approach for generating synthetic adversarial examples is to use synthetic data generation techniques. Researchers have considered using state-of-the-art patient simulators~\cite{chen_data-driven_2016,kushner_models_2019} and Generative Adversarial Networks (GANs)~\cite{baluja2017adversarial,xiao2019generating,hu2022generating} to efficiently produce perceptually realistic adversarial examples. However, both techniques have substantial limitations.
Patient simulators and GANs can struggle to fully replicate complex physiological variations found in real medical data, resulting in unrealistic examples. This difficulty arises due to the inherent complexity of human physiology and the limitations of computational modeling, which require these techniques to learn simplified models of medical scenarios. 
Adversarial examples generated by these techniques can also be biased. The quality of the adversarial examples generated by these techniques is highly dependent on the quality of the technique's training data. Thus, biases in the training data (such as under-representation of specific ethnicities) can propagate into the generated data. 
In contrast to these techniques, our approach consistently produces realistic adversarial examples by sampling them from real medical data. However, our examples may reflect the bias of the input data.

\looseness=-1
Recently, \cite{hendrycks2021natural} found that clean, realistic inputs can reliably degrade the performance of machine learning models. These inputs are referred to as \emph{natural} adversarial examples, as they are examples that occur naturally but still to lead to erroneous model predictions. In practice, natural adversarial examples are obtained by selecting them from existing datasets. Adversarial filtration is a popular approach to select natural adversarial examples from an existing dataset by removing examples that are diverse in appearance but classified easily via very predictable classification boundaries~\cite{sung1996learning}. Several works have explored filtration by removing examples solved with simple spurious cues in the image domain~\cite{hendrycks2021natural}  and natural language processing (NLP) domain~\cite{sakaguchi2021winogrande,bhagavatula2019abductive,zellers2019hellaswag,dua2019drop,bisk2020piqa,hendrycks2020aligning}. 
Unfortunately, the application of this specific method may not be suitable for time-series medical data since these datasets generally lack spurious cues. However, our approach can be considered an adversarial filtration technique that removes examples based on the lower bound of weak label confidences. We leverage the intuition that low label confidence implies a sample is more ``naturally adversarial''. An alternative way to measure label confidence is to crowdsource, \ie have multiple human labelers label an example and then compute the disagreement amongst the labelers. Hence, adversarial filtration via crowdsourced label uncertainty is an approach alternative to ours.

Adversarial examples typically have a different underlying data distribution than non-adversarial (clean) examples. This suggests that it may be possible to select natural adversarial examples from existing data by observing the input data distribution. For example, natural adversarial examples can present as outliers in the input data feature distribution~\cite{aggarwal_outlier_2013}. The feature distribution can be estimated via a density estimator, and then an outlier detection method can be used to select the natural adversarial examples. Out-of-distribution (OOD) detectors~\cite{ruff_unifying_2021} may also be used to identify natural adversarial examples in existing data. Adversarial examples are similar to the clean examples from the training data distribution but with small imperceptible perturbations, and thus can be considered OOD. This paper focuses on selecting natural adversarial examples based on label uncertainty rather than feature values. We leave the investigation of feature-based approaches to be explored in future work.

\subsection{Weakly-Supervised Data Labeling for Medical CPS} Recently, a quick and inexpensive way of labeling data has emerged, known as \emph{weakly-supervised data labeling}.  Often motivated by its need in medical applications, 
its key element 
is a set of quantitative intuitions about how the data corresponds to labels. For example, a clinician might say, ``when a patient over 60 years old has had a heart rate over 120 beats for over a minute, such an alarm is a high priority.'' These intuitions, algorithmically represented as \emph{labeling functions}, are allowed to be incomplete, sometimes incorrect, and contradictory. A labeling function returns a class label or an ``abstain'' verdict for any input. Given a diverse combination of many labeling functions and an unlabeled dataset, weakly-supervised data labeling techniques
produce probabilistic labels for each sample in the dataset in the form of probability distributions over the label space. Such a label is represented as a probability distribution over the label space. Subsequently, for each sample, the weak label consists of the label with the highest probability and the confidence equal to that probability. 
A prominent weakly-supervised data labeling technique \emph{Snorkel}~\cite{ratner2016data, ratner2020snorkel} estimates an optimal weight for each labeling function by using a generative graphical model and a prior on the class balance.
Our approach takes data and labeling functions as input, feeds them into Snorkel, and builds on the resulting probabilistic labels to order the data adversarially. 

\looseness=-1
Extending the above 
work, adversarial data programming generates data in addition to labeling it~\cite{pal2018adversarial,pal2020generative}. 
A GAN is used to estimate the weight of each labeling function as well as the dependencies between them given a set of labeling functions and an unlabeled dataset. The weights and dependencies are used by the GAN's generator to produce labeled samples that come from the data distribution. 
Hence, one may be able to train an adversarial data programming model to generate more examples given a dataset comprised of natural adversarial examples. 
However, as mentioned earlier, GAN generated examples can be unrealistic and biased -- especially in medical CPS applications where small perturbations to physiological waveforms can profoundly change their meaning.

\section{Adversarially Ordered Natural Datasets}
\label{sec:problem}


We start with several notational conventions. Given a set $B$, we write $|B|$ to be the set's cardinality. Given a value $v \in \mathbb{R}$, we write $|v|$ to be the absolute value. 
We are given an evaluation dataset $X \subset \Xs$ with \emph{unknown} true labels $Y$. 


Before formally stating the problem considered in this work, we define adversarially ordered datasets and statistically valid adversarially ordered datasets.

\begin{definition}[Adversarially Ordered Natural Datasets]\label{def:aond}
    Consider a sequence of natural (non-synthetic) datasets $D_1, \ldots, D_N$, where $D_i = (X_i, Y_i, \Yh_i)$ 
    is composed of samples $X_i$, corresponding unknown true labels $Y_i$, and corresponding known noisy labels $\Yh_i$. These datasets are \emph{adversarially ordered} if their accuracy (non-strictly) monotonically decreases. That is, 
    $$\acc(Y_1, \Yh_1) \ge \acc(Y_2, \Yh_2) \ge \ldots \ge \acc(Y_N, \Yh_N)$$ 
    where
    $$\acc(Y, \Yh) = \frac{1}{|Y|} \sum_{i=1}^{|Y|} \ind{y_i = \yh_i}$$
\end{definition}

In short, the accuracy of the weak labels for each dataset in adversarially ordered natural datasets does not increase. It is important to validate this trend to verify its direction and significance. Spearman's rank correlation coefficient $\rho$ is a widely used measure of the strength and direction of a monotonic relationship~\cite{spearman1987proof}. The coefficient $\rho$ is a value between -1 and 1, where close to -1 indicates a strong, monotonically decreasing trend and close to 1 indicates a strong, monotonically increasing trend. The coefficient is accompanied by a p-value $p^*$, which quantifies the statistical significance of the observed relationship. The following definition describes how to compute Spearman's Rank Correlation for adversarially ordered natural datasets. 

\begin{definition}[Spearman's Rank Correlation]\label{def:spearmans_rank}
    Consider adversarially ordered natural datasets $D_1, \ldots, D_N$. Let $R = (i_1, \ldots, i_N)$ where $i_n \in \{1,\dots,N\}$ be the rank order of the accuracies of the datasets (\ie $\acc(Y_{i_1}, \Yh_{i_1}) \le \acc(Y_{i_2}, \Yh_{i_2}) \le \ldots \le \acc(Y_{i_N}, \Yh_{i_N})$). 
    We compute \emph{Spearman's rank correlation} coefficient $\rho$ as follows: $$\rho = \frac{6 \sum_{j=1}^{N} (i_j - j)^2}{N (N^2 - 1)}$$ 
    The corresponding p-value is $p^* = 2 \times \P(T \ge |t|)$, where $T$ follows a t-distribution with $N - 2$ degrees of freedom and $$t = \rho \sqrt{\frac{N - 2}{1 - \rho^2}}$$ 
\end{definition}

We determine the statistical validity of adversarially ordered datasets by checking that Spearman's rank correlation is negative and p-value is below a predetermined significance level. We formalize this idea in the definition below.

\looseness=-1
\begin{definition}[Statistically Valid Adversarial Ordering] \label{def:svao}
    Consider adversarially ordered natural datasets $D_1, \ldots, D_N$. Let $\rho$ and $p^*$ be the Spearman's Rank Correlation Coefficient and corresponding p-value, computed 
    from the weak label accuracies of the datasets. The datasets are \emph{statistically valid adversarially ordered} if the coefficient is negative (\ie $\rho < 0$) and the p-value is statistically significant (\ie $p^* \le \gamma$ where $\gamma$ is the predetermined adversarial ordering significance threshold). 
\end{definition}

%


We are now ready to state our central technical problem.

\paragraph{Problem.} Given unlabeled data $X$ and labeling functions $\Lambda$, produce statistically valid adversarially ordered natural datasets $D_1, \ldots, D_N$.

\looseness=-1
Now we highlight the challenges of obtaining statistically valid adversarially ordered medical datasets --- and our steps to overcome them. 
Our main challenge is the absence of true labels in a given dataset $X$, which makes it impossible to directly compute the accuracies in Def.~\ref{def:aond}. To address this challenge, we will create a \emph{probabilistic labeler} — an algorithm that takes a sample and assigns it an estimated label and a confidence in that label (a value between 0 and 1). We refer to the assigned label as $\fh(x)$ and its confidence as $\gh(x)$ for any sample $x \in \Xs$. We will build that labeler from \emph{labeling functions}, which encode the rules of thumb and heuristics acquired from medical experts. A labeling function $\lambda : \Xs \to \Ys \cup \{0\}$ takes a sample $x \in \Xs$ and either abstains (\ie assigns label 0) or assigns one of the classes to it. Labeling functions can contradict each other or abstain in different combinations. 

Our second challenge is that probabilistic labelers are generally overconfident in their estimated labels (as our experience shows). That is, confidence scores $\gh(x)$ are unreliable and should not be used as an indication of the accuracy of label $\fh(x)$. Hence, our second task is to better quantify the uncertainty in the estimated labels. For each sample $x \in X$, we generate an interval $I$ of possible confidences in label $\fh(x)$. This interval should contain the \emph{true} confidence in label $\fh(x)$,
call it $g(x)$, with probability of at least $1 - \alpha$ where $\alpha$ is the significance level specified by the user. The next section details our solution to the problem of this paper.

\section{Dataset Curation via Adversarial Ordering}
\label{sec:methods}

\begin{figure}[t]
    \centering
    \includegraphics[width=\linewidth]{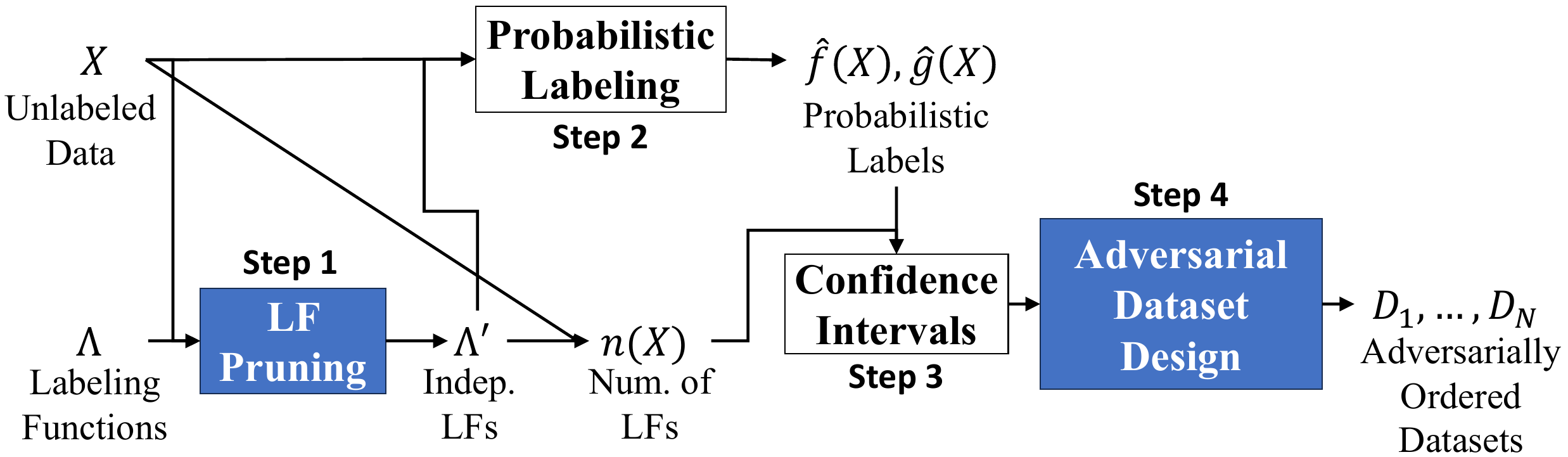}
    \caption{Our approach for curating adversarially ordered natural datasets.}
    \label{fig:approach}
\end{figure}

This section describes our approach to curate adversarially ordered natural datasets. Figure~\ref{fig:approach} 
summarizes the steps of our approach. Our approach takes as input an unlabeled dataset $X$ and a set of labeling functions $\Lambda$. The first step of our approach selects a subset of the labeling functions to be used in the next step, probabilistic labeling of the unlabeled dataset. Next, we quantify the uncertainty in the probabilistic labels by constructing confidence intervals around them. Then the confidence intervals are used in the final step to curate a sequence of progressively more adversarial datasets. 

\subsection{Labeling Function Pruning}
\label{sec:lf_select}

Labeling functions are a key component of weak supervision techniques, offering a pragmatic approach for annotating data with noisy yet informative labels. 
However, in practice, labeling functions often exhibit statistical dependencies between them, (\eg multiple functions relying on the same features or patterns)~\cite{ratner2016data}.
Dependent labeling functions result in duplicate information that can bias the outputs of weak supervision techniques, unless accounted for in the underlying model. 
Most techniques commonly assume the independence of labeling functions (conditioned on the true label); however, they also offer the option to provide labeling function dependencies as additional input~\cite{ratner2016data,ratner2019training}. 
When provided, the dependencies are embedded into the underlying model.
Unfortunately, in many applications the number of dependencies can become too large for the technique, especially when given many labeling functions.
Hence it may be preferred to provide a smaller set of independent labeling functions instead of a larger set of dependent ones.
We opt to use weakly-supervised data labeling techniques under the independent labeling function assumption.
Therefore, it is crucial to identify a subset of the labeling functions $\Lambda' \subseteq \Lambda$ where
the labeling functions are independent 
of each other prior to applying these techniques. 

\begin{algorithm}[t]
    \caption{Selecting independent labeling functions.}
    \label{alg:select_lfs}
    \begin{algorithmic}[1]
        \REQUIRE Labeling functions $\Lambda$, unlabeled data $X$, and correlation threshold $\delta$
        \ENSURE Independent subset of labeling functions \\$\Lambda' \subseteq \Lambda$
        \STATE Apply labeling functions $\Lambda$ to the data $X$ to get weak labels $\mathcal{L}_\Lambda(X)$.
        \STATE Compute the correlation $c_{ij}$ between all pairs of label- \\ing functions $\lambda_i, \lambda_j \in \Lambda$ where $i \neq j$ using $\mathcal{L}_\Lambda(X)$.
        \STATE Construct a labeling function dependency graph $G$ \\ with labeling functions $\lambda$ as nodes and edges between labeling functions (nodes) with correlation $c_{ij} > \delta$.
        \STATE Rank labeling functions in descending order by the number of maximal cliques of $G$ they belong to, breaking ties by labeling function coverage (higher coverage corresponds to a higher rank). Let $R$ denote this ranking.
        \STATE Let $\Lambda' = \Lambda$. For each labeling function $\lambda$ in the \\ ranking $R$, if $\lambda \in \Lambda'$, remove all other labeling \\ functions with which it shares a maximal clique in $G$ from $\Lambda'$.
    \end{algorithmic}
\end{algorithm}

The pseudo-code for our independent labeling function selection procedure is presented in Algorithm~\ref{alg:select_lfs}. First, the labeling functions $\Lambda$ are applied to the dataset $X$ to obtain the weak labels. We will refer to the weak labels for $X$ as $\mathcal{L}_\Lambda (X) = (\mathcal{L}_{\lambda_i} (X), \ldots, \mathcal{L}_{\lambda_{|\Lambda|}} (X))$. Next, we compute the Pearson correlation coefficients $c_{ij}$ for each pair of labeling functions $(\lambda_i, \lambda_j)$ where $\lambda_i, \lambda_j \in \Lambda$ and $i \neq j$ from the weak labels, \ie $c_{ij} = \pearson (\mathcal{L}_{\lambda_i} (X), \mathcal{L}_{\lambda_j} (X))$.
These coefficients are measures of linear correlation (dependence) between pairs of our labeling functions.

\looseness=-1
Now we aim to rank the labeling functions from most to least independent. To do this, we first construct a graph representation 
of the labeling function dependencies~\cite{ratner2019training}. 
Concretely, the nodes of the graph $G$ are the labeling functions in $\Lambda$. We add an edge to $G$ for each labeling function pair whose correlation is sufficiently large, that is, add edge $(\lambda_i, \lambda_j)$ if $|c_{ij}| > \delta$ where $\delta$ is a user-specified minimum threshold on correlation. 
Next, we identify maximal cliques of labeling functions, which effectively reveal subsets of labeling functions that tend to share similar labeling patterns (\ie cover the same information). Hence our intuition is that labeling functions belonging to many cliques are more dependent. 
We rank the labeling functions by the number of maximal cliques they belong to in $G$. Ties in the ranking are resolved by labeling function coverage, that is, the proportion of samples for which a labeling function emits a (non-abstain) label, \ie $\frac{1}{|X|} \sum_{x \in X} \ind{\lambda(x) \neq 0}$. We assign higher rank to labeling functions with higher coverage to retain a substantial proportion of weak labels for the data $X$.

Finally, we determine a subset of independent labeling functions $\Lambda' \subseteq \Lambda$. Our goal is to select the smallest subset of labeling functions that cover all the cliques. Let $\Lambda' = \Lambda$ to start. Then for each labeling function $\lambda$ in the ranking (in descending order), if $\lambda \in \Lambda'$, then remove all other labeling functions that share a maximal clique with $\lambda$. 
Now we have an independent subset  $\Lambda' \subseteq \Lambda$ for weak supervision. 

\subsection{Probabilistic Labeling}
\label{sec:prob_labeling}

In this step, we combine the weak labels produced by the labeling functions into a single ``strong'' probabilistic label. This label is characterized by a confidence score between 0 and 1, indicating the level of certainty in the label's accuracy. Mathematically, for each sample $x \in X$, we combine its weak labels $\mathcal{L}_\Lambda (x)$ into a probabilistic strong label $\fh(x)$ with confidence $\gh(x)$. 

Our approach computes the strong label and corresponding confidence using a weak supervision technique that performs a \emph{weighted combination} over the weak labels. 
The weights $w$ has a fixed vector per class containing one weight per labeling function. 
While our approach supports a variety of weighted combination techniques, we consider two of them in this paper: 
\begin{itemize}
    \item Majority vote
    \item Generative model with an uninformed prior
\end{itemize}
These two techniques are applied in two steps:
\begin{enumerate}
    \item Determine the weights $w$
    \item Combine the weak labels $\mathcal{L}_\Lambda (x)$ using weights $w$ for each $x \in X$ 
\end{enumerate}

The first step learns a non-negative weight vector where each weight indicates the relative priority of the corresponding labeling function. 
Majority vote, a widely-used and straightforward method for combining multiple discrete signals into one, assigns equal priority to each labeling function. Hence, the weight vector for majority vote is uniform (\ie 
$w_i^{(y)} = 1$ for all $i \in \{1, \ldots, |\Lambda|\}$ and $y \in \Ys$
).
Generative models are popular in state-of-the-art weak supervision literature~\cite{ratner2016data,bach2017learning,pal2018adversarial,fu2020fast}
These models give higher weights to labeling functions with higher accuracies. The accuracies are unknown a priori, so the model estimates them by observing the agreements and disagreements of the labeling functions in the data ($\mathcal{L}_\Lambda(x)$) during its training phase. Furthermore, the model can be trained with a prior that specifies the expected frequency of each label in $\Ys$. We assume the actual prior is unknown and we do not have a labeled dataset on which we can estimate it, so we use an uninformed prior which assigns equal probability to every label. The technical details on this use of generative models can be found in \cite{pugh2021high}. 

\looseness=-1
In the second step, the learned weights are used to combine the weak labels into a single probabilistic label for each sample.
Suppose now we want to label a sample $x$ given its weak labels $\mathcal{L}_\Lambda (x)$ and the weights $w$. 
First, we obtain label weights by adding up the weights of all labeling functions choosing that label, \ie
$\sum_{i=1}^{|\Lambda|} w_i^{(y)} \cdot \lambda_i (x)$ for $y \in \Ys$.
Then we pass the label weights through the softmax function, which is a standard way to normalize positive real numbers into a probability distribution. Finally, the largest normalized weight is used as the confidence $\gh(x)$ and the label corresponding to that weight becomes the estimated label $\fh(x)$. 

Unfortunately, as mentioned in the previous section, the confidence scores output by these two weak supervision techniques are typically over-confident; that is, the confidence $\gh(x)$ overestimates the accuracy of the estimated label $\fh(x)$. In other words, weak supervision techniques are poorly calibrated~\cite{guo_calibration_2017} -- the confidence scores do not reflect the true accuracy of the label. Majority vote can be especially prone to over-confidence because if many of the labeling functions are inaccurate, the label prediction can be incorrect with high confidence. Generative models have an advantage over majority vote because they can account for the inaccuracies of labeling functions and, thus, yield more accurate confidences. However, as we witness in practice, generative models are also often poorly calibrated. 
Hence, the confidences output by weak supervision techniques should be used with caution or disregarded entirely.
In response, we propose using confidence intervals to account for the uncertainty in these confidences.

\subsection{Confidence Intervals for Weak Label Accuracies}
\label{sec:conf_intervals}

Our confidences in the estimated labels from the prior subsection can be unreliable. More precisely, the confidences in the estimated labels may not reflect the true accuracy of those estimated labels. We quantify the potential inaccuracy in the estimated labels by providing \emph{confidence intervals}. 
Our confidence intervals contain the likely true confidences in the estimated labels. For each sample $x \in X$, we aim to generate an interval $I$ containing the true confidence in the estimated label $\fh(x)$, call it $g(x)$, with probability of at least $1 - \alpha$, \ie $\P[g(x) \in I] \ge 1 - \alpha$.

We bound the unknown true confidences $g(X)$ in our estimated labels $\fh(X)$ using the Clopper-Pearson (CP) interval $[\theta_L, \theta_U]$ where $\theta_L, \theta_U \in [0,1]$ and $\theta_L < \theta_U$~\cite{clopper1934use}. 
This interval bounds the true success probability $\mu$, constructed from a sample $s \sim B(n, \mu)$ from a binomial distribution with $n$ trials and success probability $\mu$, which holds with probability at least $1 - \alpha$, \ie $\P_{s \sim B(n, \mu)} \left[ \theta_L (\alpha; n, s) \le \mu \le \theta_U (\alpha; n, s) \right] \ge 1 - \alpha$. 

Intuitively, few non-abstaining labeling functions with large weights should yield a lower confidence than many non-abstaining labeling functions with moderate weights. 
Hence, in addition to the labeling function weights, the \emph{number of labeling functions} $n$ contributing to the estimation of the labels should be considered in the calculation of confidence in $\fh(x)$.
While the confidences provided by probabilistic labelers $\gh(X)$ do not take this into account (which can lead to their unreliability), we incorporate the number of voting labeling functions into the construction of our intervals.
When a labeling function emits a (non-abstain) label for some sample $x$, we consider it a Bernoulli trial whose outcome is a success if that label is correct or a failure if it is incorrect. 
We let the number of successes $s$ (of the $n$ trials) be 
the normalized probability of label $\fh(x)$
weighted by the number of non-abstaining labeling functions $n$.
Hence by definition of the CP interval, we derive the interval $I = [\theta_L, \theta_U]$ for $g(x)$ given any sample $x \in X$ and significance level $\alpha$ as,
$$\theta_L (\alpha; n, s) = B \left( \frac{\alpha}{2}; s, \left( n - s + 1 \right) \right)$$
$$\theta_U (\alpha; n, s) = B \left( 1 - \frac{\alpha}{2}; (s + 1), \left( n - s \right) \right)$$
where 
%
$$n(x) = \sum_{i=1}^{|\Lambda|} \ind{\lambda_i (x) \neq 0} $$
$$s(x) = n(x) \cdot \frac{\mbox{exp}\left\{ \sum_{i=1}^{|\Lambda|} w_i^{\left( \fh(x) \right)} \lambda_i (x) \right\}}{\sum_{y \in \Ys} \mbox{exp}\left\{ \sum_{i=1}^{|\Lambda|} w_i^{(y)} \lambda_i (x) \right\}}$$
%
where 
$B(q; a, b)$ is the $q$-th quantile from a beta distribution with shape parameters $a$ and $b$.

\subsection{Adversarial Dataset Curation}
\label{sec:dataset_design}

Now we create a sequence of adversarially ordered natural datasets. Our confidence interval lower bounds indicate the smallest possible certainty in the estimated label. So as more samples with small lower bounds are included in a dataset, the more adversarial it gets. We arrange the samples in $X$ by their confidence interval lower bound in descending order, \ie $x_{i_1}, x_{i_2}, \ldots, x_{i_{|X|}}$ 
where $\theta_L(x_{i_1}) \ge \theta_L(x_{i_2}) \ge \ldots \ge \theta_L(x_{i_{|X|}})$ and $i_1, \ldots, i_{|X|} \in \{1, \ldots, |X|\}$. Then we produce a sequence of datasets $D_1, \ldots, D_N$ where each dataset $D_n$ contains the top $\frac{i}{N}$ percent of ordered samples and their corresponding estimated labels, that is, $$D_n = \{(x_{i_j}, \fh(x_{i_j}))\} \quad \mbox{for } j \in \left\{1, \ldots, \frac{i \cdot |X|}{N} \right\}$$ 
In conclusion, we have natural datasets $D_1, \ldots, D_N$ that are progressively more adversarial.

\section{Results}
\label{sec:results}

In this section, we evaluate our approach on nine case studies (six with clinical relevance). Specifically, we summarize the nine case studies 
describe comparative approaches, 
and evaluate the adversarial ordering of natural adversarial datasets produced by our approach. 

\subsection{Case Studies}
\label{sec:examples}

We evaluate our approach on five physiological alarm suppression case studies, one clinical text classification case study, and three non-clinical text classification case studies. These datasets are summarized in Table~\ref{tab:examples}.

\begin{table}[t]
    \centering
    \begin{tabular}{lcccc}
        \toprule
        Example & Train & Valid & Test & Num. LFs \\
        \midrule
        HR Low & 79 & - & - & 62 \\
        HR High & 1315 & - & - & 62 \\
        RR Low & 312 & - & - & 62 \\
        RR High & 574 & - & - & 62 \\
        \spotwo Low & 3265 & - & - & 62 \\
        Crossmodal & 2630 & 376 & 378 & 18 \\
        Crowdsourcing & 187 & 50 & 50 & 103 \\
        Recsys & 796956 & 8339 & 42191 & 5 \\
        Spam & 1586 & - & 250 & 9 \\
        \bottomrule 
    \end{tabular}
    \caption{Summary of our nine evaluation datasets. 
    }
    \label{tab:examples}
\end{table}

\paragraph{Physiologic Alarm Suppression} Alarm suppression, which involves distinguishing suppressible (uninformative to clinical care) and non-suppressible physiologic-monitoring alarms, is the aim of the first five datasets. Datasets of heart rate (HR) low/high, respiratory rate (RR) low/high, and \spotwo low alarms were extracted from~\cite{macmurchy2017acceptability}. Each alarm sample is represented as a multi-vital sign time series data. We use the set of sixty-two clinician-designed labeling functions developed for this example from~\cite{pugh2022evaluating, pugh2021high}. The labeling functions analyze the time series data to make predictions on suppressibility, \eg an alarm is non-suppressible if the heart rate is above 220 for longer than 10 seconds after the alarm starts, otherwise it abstains.

\paragraph{Clinical Text Classification} The \textbf{Crossmodal} dataset aims to label radiography, computed tomography, and electroencephalography images by writing labeling functions over an auxiliary modality, namely, corresponding imaging text reports~\cite{dunnmon2019crossmodal}. The labeling functions in this example are clinician-designed, expressing simple pattern-matching or ontology-lookup heuristics.

\paragraph{Non-Clinical Text Classification} The \textbf{Crowdsourcing}, \textbf{Recsys}, and \textbf{Spam} datasets are publicly-available at www.snorkel.org. 
The objective of \textbf{Crowdsourcing} is to label tweets pertaining to weather expressing either a positive or negative sentiment. 
\textbf{Recsys} aims to predict whether a user will read and like any given book or not. 
Finally, the \textbf{Spam} dataset aims to classify spam emails. 

\subsection{Implementation}
\label{sec:implementation}

In this section, we provide details on the implementation of our approach for curating adversarially ordered natural datasets. The implementation of probabilistic labeling via majority vote is straightforward. For probabilistic labeling via a generative model, we use a tool called Snorkel to train a generative model. 
Snorkel is the state-of-the-art tool for weak label combination and has been applied to several applications. We use the latest version at the time of writing, version 0.9.7 (www.snorkel.org).

For our confidence intervals, we use the confidence interval for a binomial proportion implementation, namely the ``proportion\_confint'' function, from the statsmodels Python library, version 0.14.0 (www.statsmodels.org). We specify the hyperparameters such that the Clopper-Pearson interval based on the Beta distribution with a 5\% significance level is used. Specifically, hyperparameters ``alpha'' and ``method'' are set to 0.05 and ``beta'', respectively.

Generative models do not generalize to unseen samples, so we combine the train, validation, and test splits (without labels) as training data for the generative model. In order to evaluate the accuracy of the noisy datasets produced by our approach, we select the samples to be included in our datasets from the available labeled data (\ie train for the alarm suppression examples and the validation and test splits combined for all other examples).

\subsection{Comparative Approaches}
\label{sec:comparative}

We previously discussed two primary challenges of probabilistic labeling via weak supervision techniques: (1) labeling function dependence, and (2) unreliability of confidence scores output by probabilistic labelers. In our evaluation, we will demonstrate why addressing these challenges is necessary to achieve sufficient performance. Hence we define three comparative approaches as follows. \textbf{PL Conf with all LFs} 
produces datasets by ordering samples by the probabilistic labeler's confidences ($\gh(X)$) computed using \emph{all} of the provided labeling functions (regardless of the dependencies between them), and then selecting the top-$p$ percent of samples per dataset (as done in our approach). \textbf{PL Conf with Indep. LFs} repeats the same steps as \textbf{PL Conf with all LFs} but uses a set of independent labeling functions selected by the same procedure from our approach. 
Lastly, \textbf{CI LB w/ all LFs} is an instance of our approach that skips the labeling function pruning step, 
that is, it is our approach using \emph{all} provided labeling functions.

\subsection{Evaluation of Natural Adversarial Ordering}
\label{sec:evaluation}


\begin{table*}[ht!]
    \centering
    \begin{tabular}{lllcccc}
        \toprule
        \multirow{2}{*}{Study Type}  & \multirow{2}{*}{Case Study} & Probabilistic & \multicolumn{1}{c}{PL Conf} & \multicolumn{1}{c}{PL Conf} & \multicolumn{1}{c}{CI LB} & \multicolumn{1}{c}{Our} \\
        & & Labeler (PL) & \multicolumn{1}{c}{with all LFs} & \multicolumn{1}{c}{with Indep. LFs} & \multicolumn{1}{c}{with all LFs} & \multicolumn{1}{c}{Approach} \\
        \midrule
        Medical & HR Low & Majority Vote & -0.719 & -0.903 & -0.863 & -0.730 \\
        & & Snorkel & -- & -- & -0.827 & -- \\
        & HR High & Majority Vote & {\color{red} 0.818} & -0.976 & {\color{red} 0.857} & -1.000 \\
        & & Snorkel & {\color{red} 0.964} & {\color{red} 0.927} & {\color{red} 0.988} & -0.806 \\
        & RR Low & Majority Vote & -0.997 & -0.879 & -0.997 & -0.891 \\
        & & Snorkel & -0.997 & -- & -0.997 & -0.806 \\
        & RR High & Majority Vote & -1.000 & -0.988 & -1.000 & -0.952 \\
        & & Snorkel & -1.000 & -0.952 & -1.000 & -- \\
        & \spotwo Low & Majority Vote & -- & {\color{red} 0.891} & -- & -- \\
        & & Snorkel & -0.988 & -- & -- & -- \\
        & Cross Modal & Majority Vote & -1.000 & -1.000 & -0.879 & -0.782 \\
        & & Snorkel & -1.000 & -0.988 & -- & -0.806 \\
        \midrule
        Non-medical & Crowdsourcing & Majority Vote & -0.864 & -0.988 & -0.864 & -0.976 \\
        & & Snorkel & -0.976 & -0.988 & -0.864 & -0.976 \\
        & Recsys & Majority Vote & -1.000 & -1.000 & -- & -- \\
        & & Snorkel & -0.988 & -0.988 & -- & -- \\
        & Spam & Majority Vote & -0.985 & -0.985 & -0.767 & -0.767 \\
        & & Snorkel & -0.673 & -0.673 & -0.656 & -0.656 \\
        \bottomrule
    \end{tabular}
    \caption{Spearman's Rank Correlation coefficients of our adversarially ordered natural datasets}
    \label{tab:dataset_consistency}
\end{table*}


Finally, we present the results of our approach applied to several case studies. Recall that the goal of the paper is to generate a sequence of statistically valid adversarially ordered natural datasets as per Def.~\ref{def:svao}. We validate the progressive increase of ``adversarialness'' in the datasets by analyzing the Spearman's rank correlation (Def.~\ref{def:spearmans_rank}) of the adversarially ordered datasets. 

The parameters of our approach are set as follows. Correlation between two labeling functions greater than 0.5 indicates that the labeling functions are dependent, \ie correlation threshold $\delta = 0.5$. We allow for a 5\% chance of the confidence intervals not containing the true confidence in our probabilistic labels, \ie significance level $\alpha = 0.05$. Lastly, we let the, \ie adversarial ordering significance threshold $\gamma = 0.05$. 
We report the Spearman's rank correlation coefficients and corresponding p-values for the accuracies of the datasets produced by our approach and by the comparative approaches in Table~\ref{tab:dataset_consistency}. We plot the actual accuracies of the datasets in the Appendix. 

\textbf{Main Takeaway:
The results in Table~\ref{tab:dataset_consistency} demonstrate that, unlike other approaches, our approach did not yield a statistically valid non-adversarially ordered dataset using real-world medical data.}  The negative correlation coefficients in Table~\ref{tab:dataset_consistency} correspond to adversarially ordered datasets that have (generally) decreasing accuracy of weak labels.  We recall that the central premise of this work is that statistically decreasing accuracy must be assured for any adversarially ordered dataset since decreasing accuracy across the ordered datasets is the property that will ultimately be tested as a measure of robustness.  Simply put, if an approach yields an positive correlation coefficient in any application, it is unattractive as a method of curating adversarially ordered datasets.
The results demonstrate that our approach yields statistically significant adversarially ordered datasets in twelve cases, and statistically invalid datasets on all other examples. We interpret this as a substantive outcome: our approach is applicable in a majority of the examples, and never yields datasets with statistically valid increasing accuracy. All the comparative approaches produce at least as many datasets with statistically valid decreasing accuracy, but also produce datasets with statistically significant monotonically increasing accuracy --- thus violating our requirements. If used for robustness evaluation of learning models in practice, such datasets can cause incorrect and misleading results. Hence, we conclude that our approach yields more reliable results than the comparative approaches.

\section{Discussion}
\label{sec:discussion}

In this discussion we review the impact and methods for improving the curation of adversarial ordered datasets and provide a discussion of the limitations of this technique.

\subsection{Adversarial datasets need not be perfectly ordered.}

Our approach aims to construct adversarially ordered natural datasets where, by construction, the label accuracy of each dataset progressively decreases. However, perfect ordering (\ie a Spearman's Rank Correlation coefficient of exactly -1) is not required to evaluate robustness on our datasets; a statistically valid decreasing trend is sufficient (\ie coefficient is negative and p-value is sufficiently low). When evaluating a learning-enabled component, we will observe the trend in its accuracy on these datasets and test that the trend is both decreasing and significant. This pragmatic approach aligns with the practical nature of real-world data, where inherent complexities and uncertainties may lead to variations in label accuracy. Thus, by focusing on the presence of a significant negative trend, we account for the genuine challenges posed by naturally occurring adversarial examples, making our evaluation more realistic and applicable to real-world scenarios.

\subsection{Improving significance via data availability and weak labeling functions.}

To comprehensively explore the real-world applicability of our method for curating adversarially ordered natural datasets, it is important to understand the contexts under which our method is ideally suited. Recall that our method constructs its datasets by sampling from the unlabeled input data. Consequently, the quantity and quality of the input data can have a significant impact on the method's efficacy. A larger input dataset, for example, typically provides a more diverse and representative sample of the underlying data distribution. As a result, it can include a broader set of natural adversarial examples that our method can use. Additionally, recall that our method relies heavily on probabilistic labels produced by weakly-supervised data labeling techniques to identify the natural adversarial examples. Larger input data typically improves the accuracy of weakly-supervised data labeling techniques, and consequently the probabilistic labels. Since our method relies heavily on these labels, improving their accuracy can directly improve the quality of our datasets.  However, as mentioned previously, data quality is also an important factor. Even when given more input data, such data should have limited bias and errors.

Labeling function quality is also an important factor for our method's applicability. Weak labels output by labeling functions are the foundation upon which weakly-supervised data labeling techniques learn to how to label data. Hence low-accuracy labeling functions can lead to inaccurate probabilistic labels, which is likely to cause our approach to construct improper datasets. A common assumption of weakly-supervised data labeling techniques are that the provided labeling functions are at least 50\% accurate, but higher accuracy is generally more desirable. Unfortunately, how to write quality labeling functions for weak supervision is an open area research area~\cite{varma2018snuba,das2020goggles}. 
However, accurate labeling function design is not within the scope of this paper thus we assume the engineer has ensured the quality of the supplied labeling functions.

In summary, the curation of adversarial ordered datasets can be further improved by:
\begin{itemize}

\item increasing the amount of unlabeled examples

\item increasing the number and accuracy of weak labeling functions

\end{itemize}

\subsection{Significance detection remains an open challenge.}

As observed in Section~\ref{sec:results}, our approach can yield adversarially ordered natural datasets with statistically insignificant ordering. Such datasets should not be used for robustness analysis as it is unclear what the expected trend of a learning-enabled component's accuracy should be on datasets with insignificant adversarial ordering. 
For the evaluation in this paper, we determined statistical significance of our datasets by leveraging the ground-truth labels of the input data. Ground-truth labels are important for determining the expected output, the basis of a comparison to the observed outcomes to quantify significance. Our high-level goal is to provide a method for evaluating robustness to natural adversarial examples when ground-truth labels are unavailable or prohibitively expensive to obtain. Unfortunately, the development of methods for testing statistical significance in the absence of ground truth labels is an open area of research. We plan to explore this in future work.

\subsection{Potential Commercial Value.}

\looseness=-1
This work presents a method for curating naturally adversarial ordered datasets for learning-enabled medical CPS.  But, the fact remains that the gold standard for evaluating safety and efficacy of learning-enabled medical CPS is a clinical trial. Unfortunately, one of the most expense aspects of medical CPS development is real-world experimentation (i.e., observational/clinical trials) -- often requiring years of engineering development and obtaining regulatory approvals to execute.  Different from traditional labeled adversarial dataset robustness analysis, the approach presented herein is light weight -- only requiring access to unlabeled (previously collected) examples and weak labeling functions.  This is a commercial asset of the proposed approach to early stage development of learning-enabled medical CPS technologies (i.e., before labeled observational data collection).  Leveraging the work herein, it is now feasible to evaluate the robustness in several commercially important scenarios.  For example, an unlabeled dataset with different demographics than the training set could be used to assess whether a learning-enable medical CPS exhibits inherent demographic bias -- a major issue in modern medical technology development.  Another example is the proposed technique can be used in coordination with unsupervised techniques to assess real-world robustness in the absence of labeled data.  Lastly, in the commercial development of learning-enabled medical CPS technologies -- due to the pressures of raising capital and healthcare economics -- it is important to "fail fast" (i.e., to quickly rule out ideas that are unlikely to succeed) prior to incurring significant development costs.  This work provides a technique that may help early-stage researchers and entrepreneurs alike identify foundational robustness flaws in their approach prior to expensive data collection.

\section{Conclusion}
\label{sec:conclusion}

In this paper, we proposed an approach for curating a sequence of adversarially ordered natural datasets for the purpose of evaluating model robustness to natural adversarial examples. Our approach identifies a set of independent labeling functions to use for probabilistic labeling. Probabilistic labels obtained via weak supervision techniques were used as proxy of the unknown true labels. We quantify the uncertainty in these labels with Clopper-Pearson confidence bounds, and construct our datasets according to the lower bound which is a indication of how ``naturally adversarial'' a sample may be. Finally, we evaluated our approach on six clinical case studies and three others and showed that we successfully produce natural datasets with statistically valid adversarial ordering, and do not produce datasets with statistically valid non-adversarial ordering.
Directions for future work include (1) evaluating the robustness of real classifiers using our statistically valid adversarially ordered natural datasets, (2) devising a significance detector for adversarial ordering, (3) developing weakly-supervised methods for evaluating additional properties of deep learning models, and (4) calibrating weakly-supervised data labeling techniques.

\section*{Acknowledgment}
This work was supported in part by NSF-1915398 and NIH R18 HS026620.
\bibliographystyle{IEEEtran}
\bibliography{IEEEabrv,references}
%



\begin{table*}[t!]
    \centering
    \begin{tabular}{llrrrr}
        \toprule
        \multirow{2}{*}{Case Study} & Probabilistic & \multicolumn{1}{c}{PL Conf} & \multicolumn{1}{c}{PL Conf} & \multicolumn{1}{c}{CI LB} & \multicolumn{1}{c}{Our} \\
        & Labeler (PL) & \multicolumn{1}{c}{with all LFs} & \multicolumn{1}{c}{with Indep. LFs} & \multicolumn{1}{c}{with all LFs} & \multicolumn{1}{c}{Approach} \\
        \midrule
        HR Low & Majority Vote & {\color{mygreen} -0.719 (0.019)} & {\color{mygreen} -0.903 (0.000)} & {\color{mygreen} -0.863 (0.001)} & {\color{mygreen} -0.730 (0.017)} \\
        & Snorkel & -0.313 (0.379) & -0.730 (0.017) & {\color{mygreen} -0.827 (0.003)} & 0.595 (0.070) \\
        HR High & Majority Vote & {\color{red} {\color{red} 0.818 (0.004)}} & {\color{mygreen} -0.976 (0.000)} & {\color{red} 0.857 (0.002)} & {\color{mygreen} -1.000 (0.000)} \\
        & Snorkel & {\color{red} 0.964 (0.000)} & {\color{red} 0.927 (0.000)} & {\color{red} 0.988 (0.000)} & {\color{mygreen} -0.806 (0.005)} \\
        RR Low & Majority Vote & {\color{mygreen} -0.997 (0.000)} & {\color{mygreen} -0.879 (0.001)} & {\color{mygreen} -0.997 (0.000)} & {\color{mygreen} -0.891 (0.001)} \\
        & Snorkel & {\color{mygreen} -0.997 (0.000)} & -0.394 (0.260) & {\color{mygreen} -0.997 (0.000)} & {\color{mygreen} -0.806 (0.005)} \\
        RR High & Majority Vote & {\color{mygreen} -1.000 (0.000)} & {\color{mygreen} -0.988 (0.000)} & {\color{mygreen} -1.000 (0.000)} & {\color{mygreen} -0.952 (0.000)} \\
        & Snorkel & {\color{mygreen} -1.000 (0.000)} & {\color{mygreen} -0.952 (0.000)} & {\color{mygreen} -1.000 (0.000)} & -0.455 (0.187) \\
        \spotwo Low & Majority Vote & -0.309 (0.385) & {\color{red} 0.891 (0.001)} & -0.248 (0.489) & 0.188 (0.603) \\
        & Snorkel & {\color{mygreen} -0.988 (0.000)} & 0.430 (0.214) & -0.455 (0.187) & -0.612 (0.060) \\
        Cross Modal & Majority Vote & {\color{mygreen} -1.000 (0.000)} & {\color{mygreen} -1.000 (0.000)} & {\color{mygreen} -0.879 (0.001)} & {\color{mygreen} -0.782 (0.008)} \\
        & Snorkel & {\color{mygreen} -1.000 (0.000)} & {\color{mygreen} -0.988 (0.000)} & -0.624 (0.054) & {\color{mygreen} -0.806 (0.005)} \\
        Crowdsourcing & Majority Vote & {\color{mygreen} -0.864 (0.001)} & {\color{mygreen} -0.988 (0.000)} & {\color{mygreen} -0.864 (0.001)} & {\color{mygreen} -0.976 (0.000)} \\
        & Snorkel & {\color{mygreen} -0.976 (0.000)} & {\color{mygreen} -0.988 (0.000)} & {\color{mygreen} -0.864 (0.001)} & {\color{mygreen} -0.976 (0.000)} \\
        Recsys & Majority Vote & {\color{mygreen} -1.000 (0.000)} & {\color{mygreen} -1.000 (0.000)} & -0.321 (0.365) & -0.321 (0.365) \\
        & Snorkel & {\color{mygreen} -0.988 (0.000)} & {\color{mygreen} -0.988 (0.000)} & -0.539 (0.108) & -0.539 (0.108) \\
        Spam & Majority Vote & {\color{mygreen} -0.985 (0.000)} & {\color{mygreen} -0.985 (0.000)} & {\color{mygreen} -0.767 (0.010)} & {\color{mygreen} -0.767 (0.010)} \\
        & Snorkel & {\color{mygreen} -0.673 (0.033)} & {\color{mygreen} -0.673 (0.033)} & {\color{mygreen} -0.656 (0.039)} & {\color{mygreen} -0.656 (0.039)} \\
        \bottomrule
    \end{tabular}
    \caption{Spearman's Rank Correlation of the adversarially ordered natural datasets. The values reported are the correlation coefficient and corresponding p-value in parentheses. We desire datasets with statistically significant monotonically decrease. Hence we apply a 5\% significance threshold to the p-values, and color statistically significant and negative coefficients in green and color statistically significant and positive coefficients in red.}
    \label{tab:apdx_rankcors}
\end{table*}

\appendix

\subsection{Additional Results}
\label{sec:results_ctd}

\begin{figure*}[ht]
    \centering
    \subfigure{
        \includegraphics[width=0.22\linewidth]{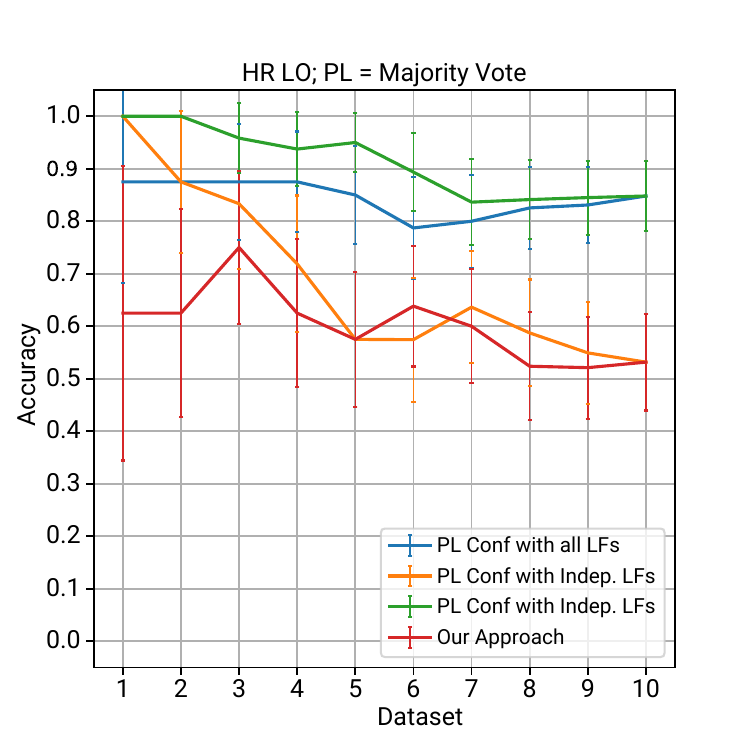}
    }
    \hfill
    \subfigure{
        \includegraphics[width=0.22\linewidth]{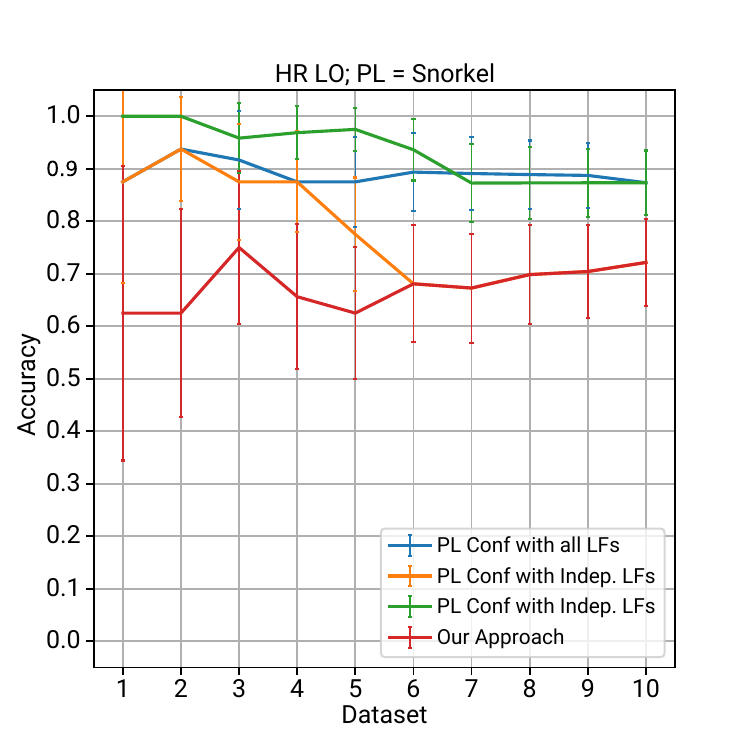}
    }
    \hfill
    \subfigure{
        \includegraphics[width=0.22\linewidth]{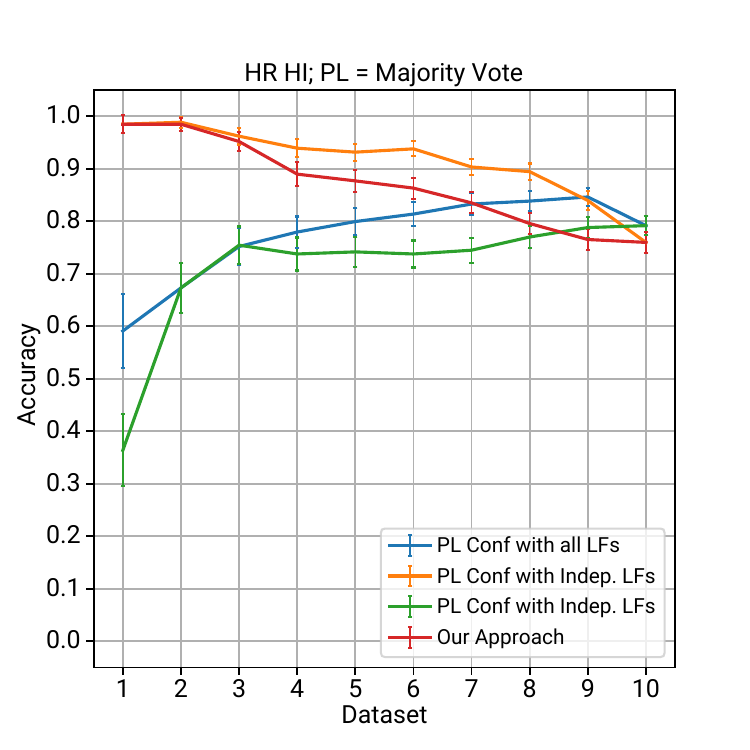}
    }
    \hfill
    \subfigure{
        \includegraphics[width=0.22\linewidth]{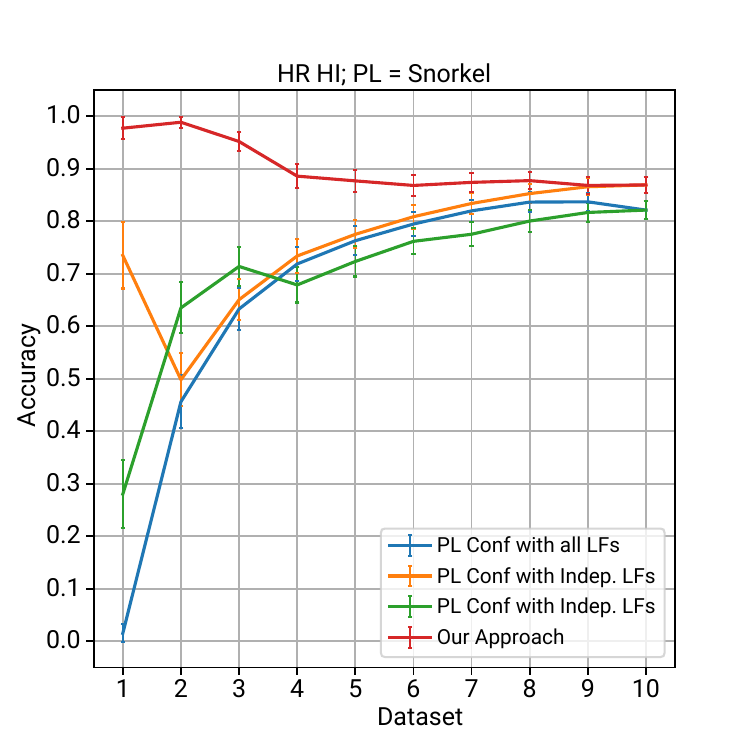}
    }
    \subfigure{
        \includegraphics[width=0.22\linewidth]{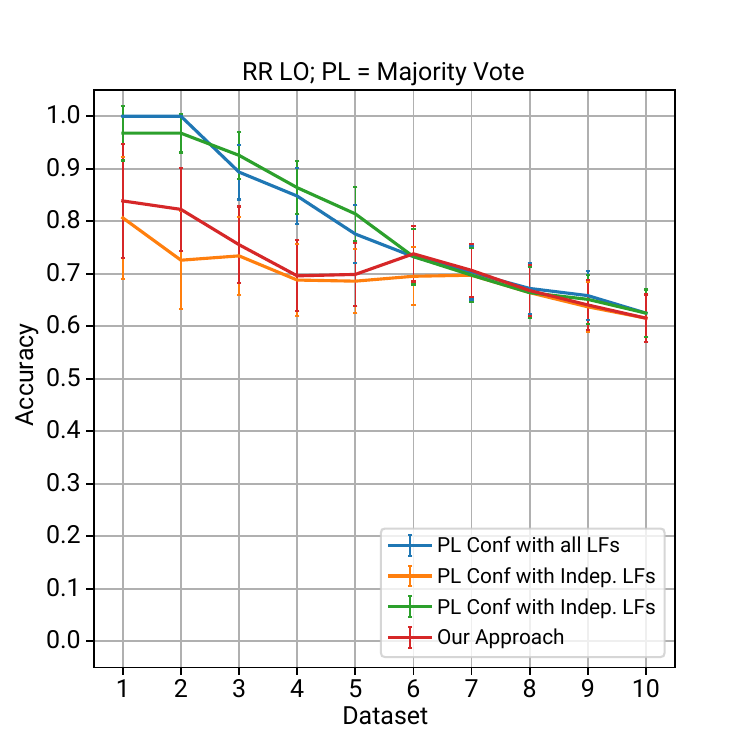}
    }
    \hfill
    \subfigure{
        \includegraphics[width=0.22\linewidth]{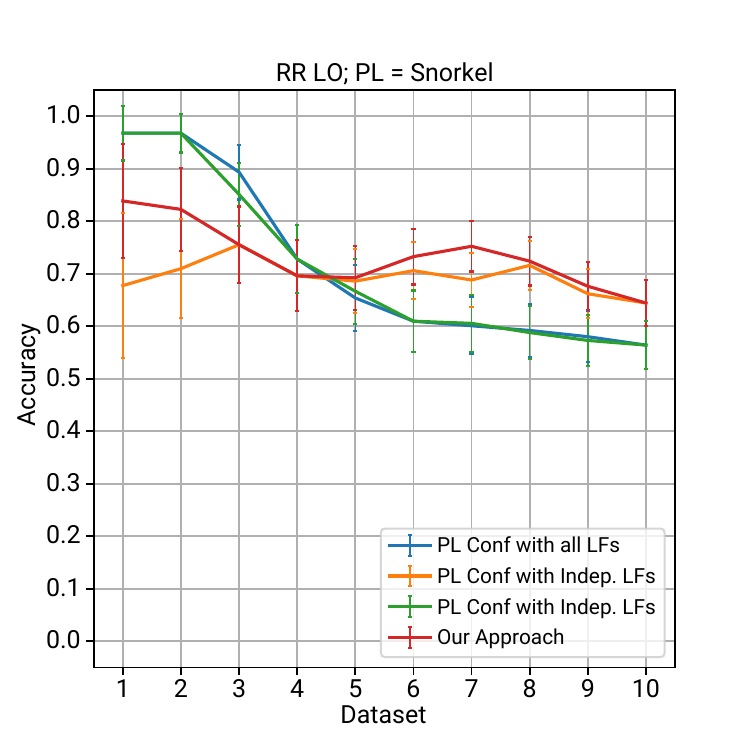}
    }
    \hfill
    \subfigure{
        \includegraphics[width=0.22\linewidth]{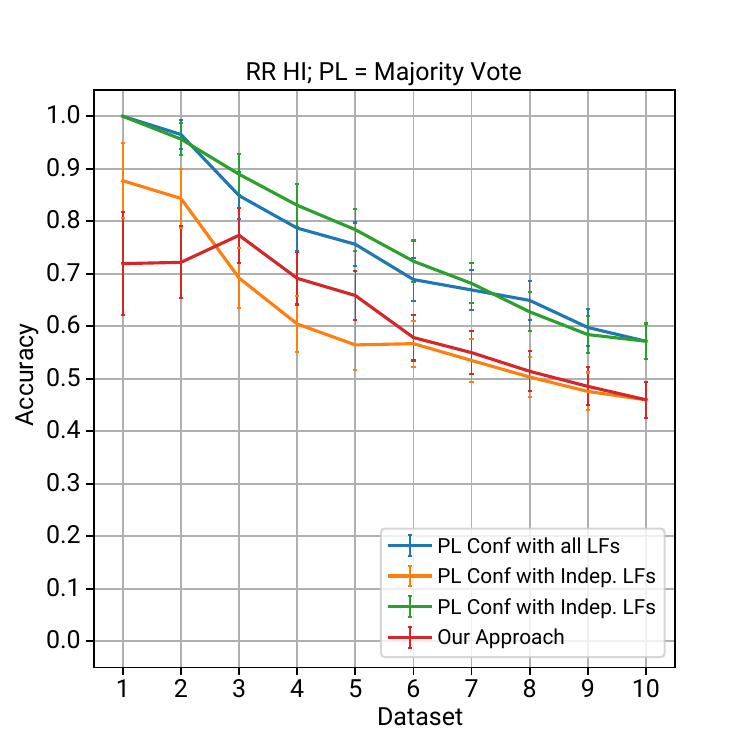}
    }
    \hfill
    \subfigure{
        \includegraphics[width=0.22\linewidth]{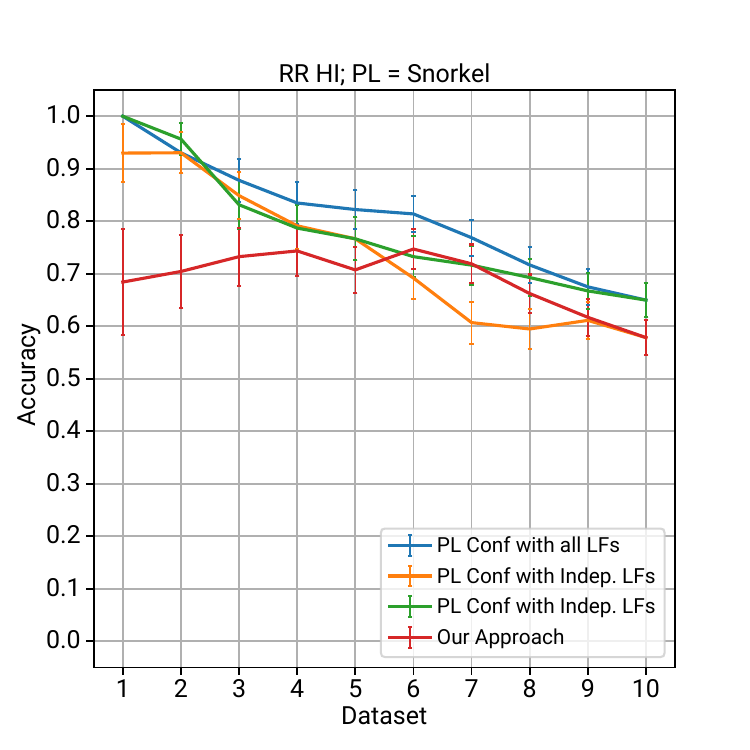}
    }
    \subfigure{
        \includegraphics[width=0.22\linewidth]{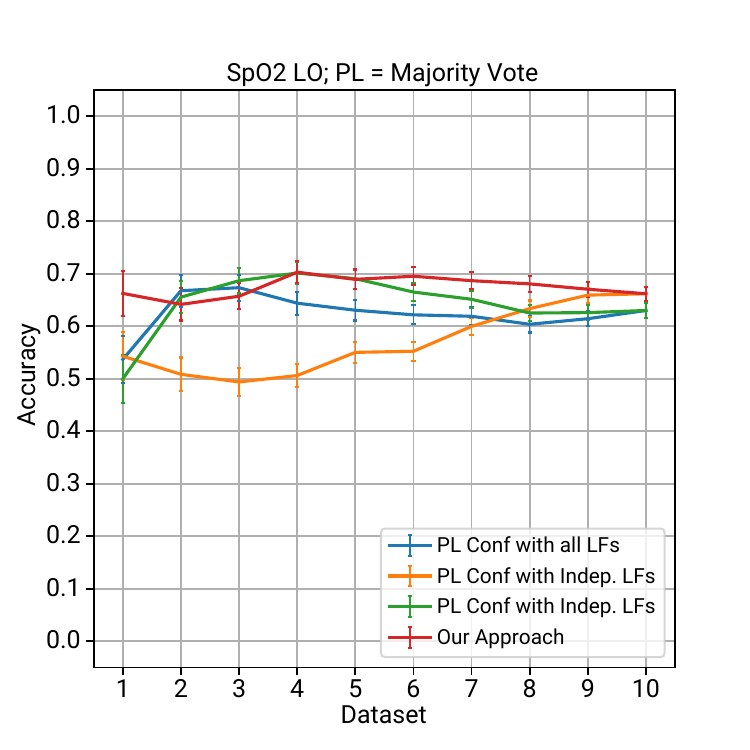}
    }
    \hfill
    \subfigure{
        \includegraphics[width=0.22\linewidth]{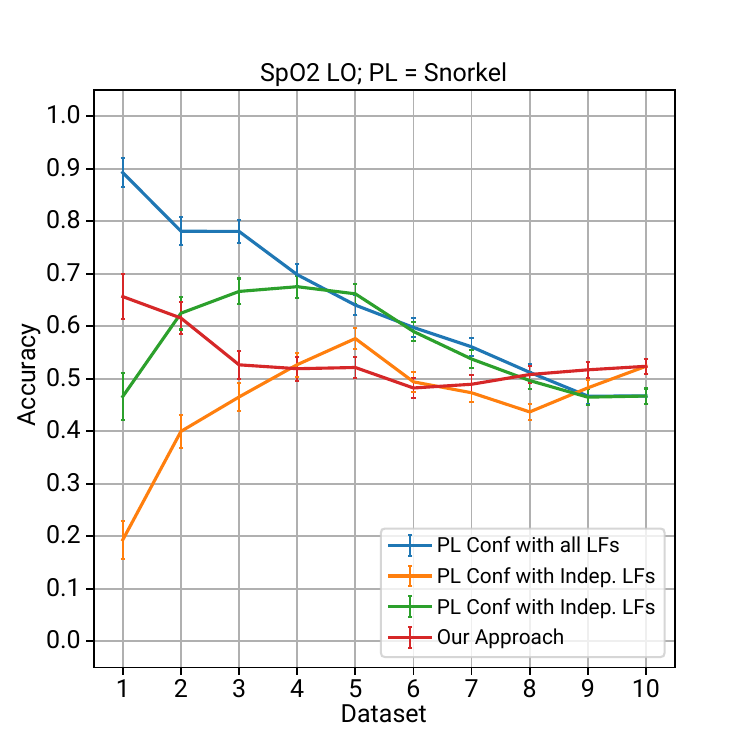}
    }
    \hfill
    \subfigure{
        \includegraphics[width=0.22\linewidth]{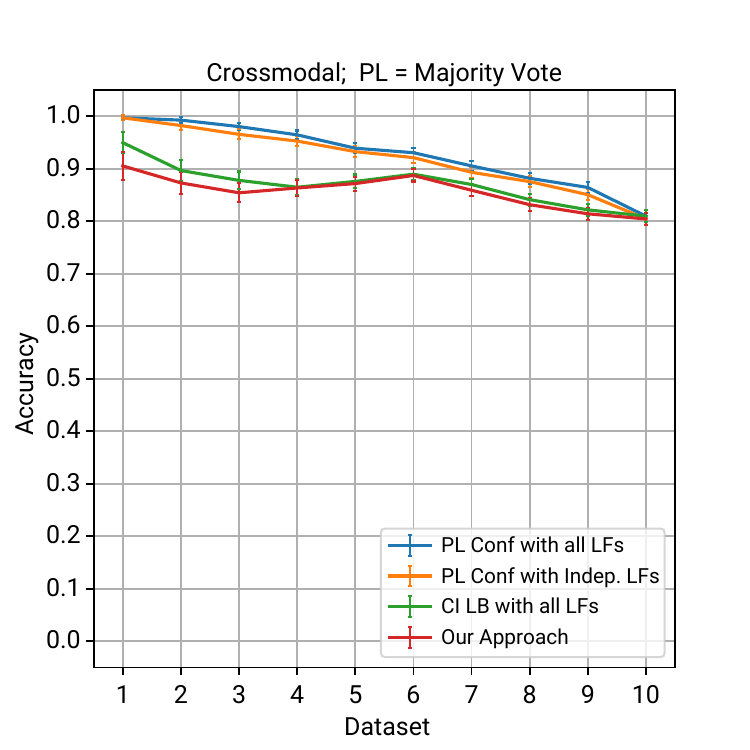}
    }
    \hfill
    \subfigure{
        \includegraphics[width=0.22\linewidth]{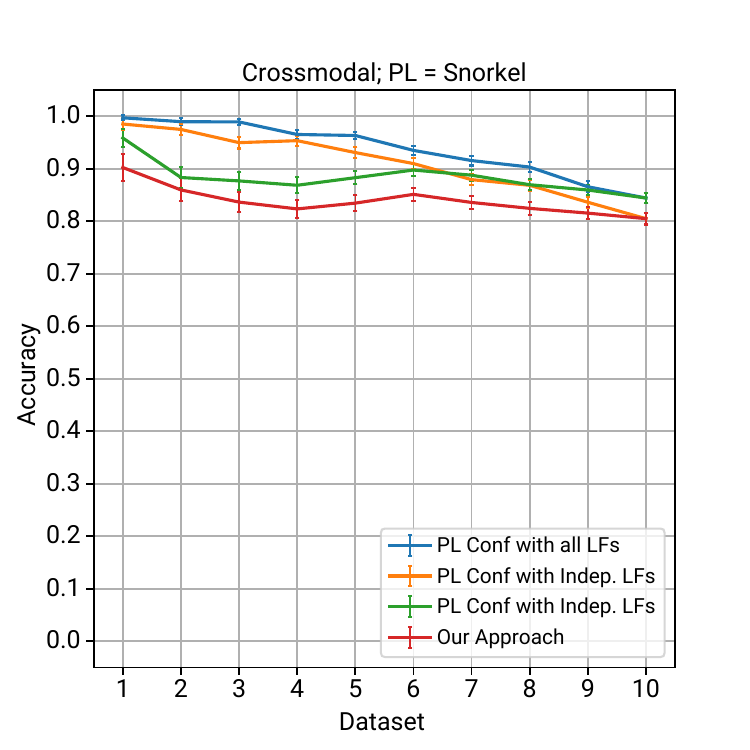}
    }
    \subfigure{
        \includegraphics[width=0.22\linewidth]{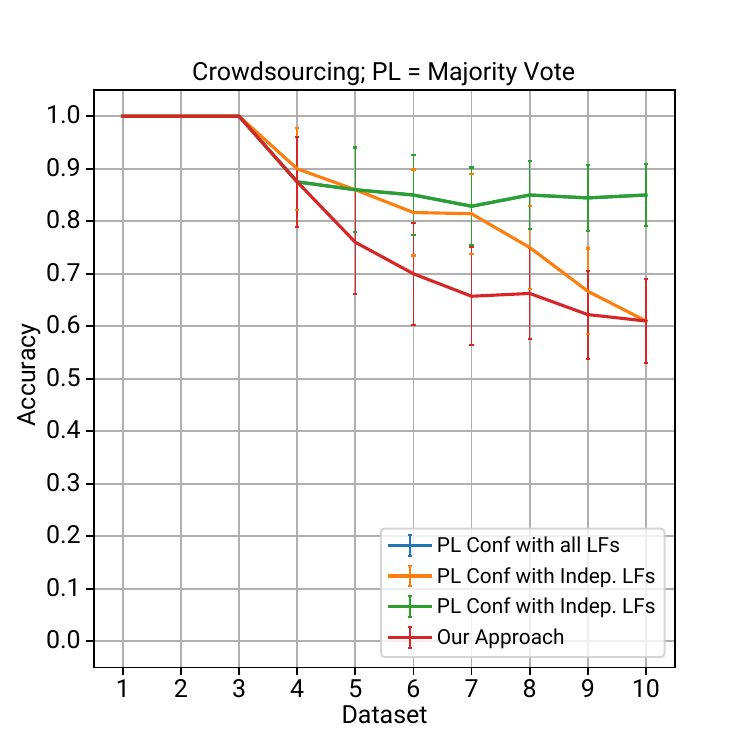}
    }
    \hfill
    \subfigure{
        \includegraphics[width=0.22\linewidth]{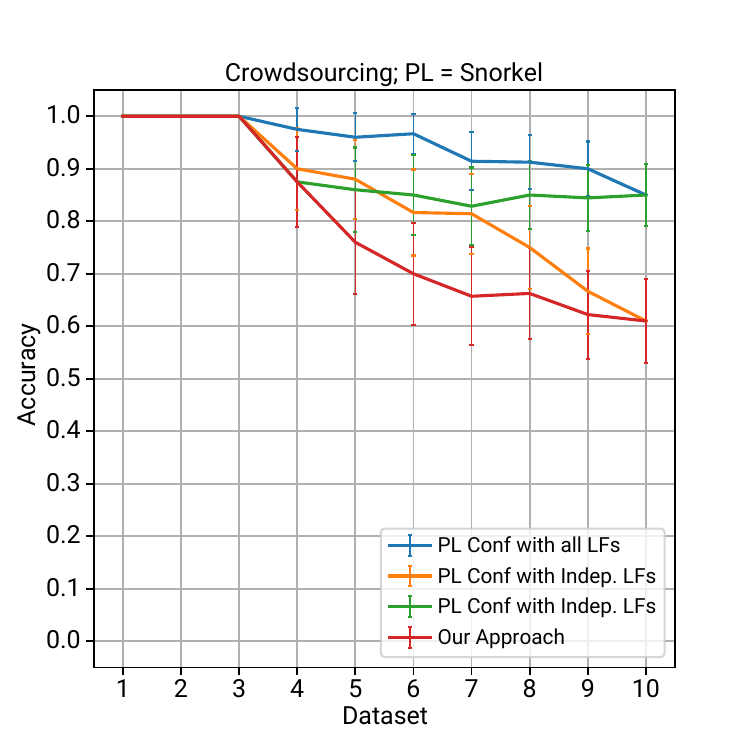}
    }
    \hfill
    \subfigure{
        \includegraphics[width=0.22\linewidth]{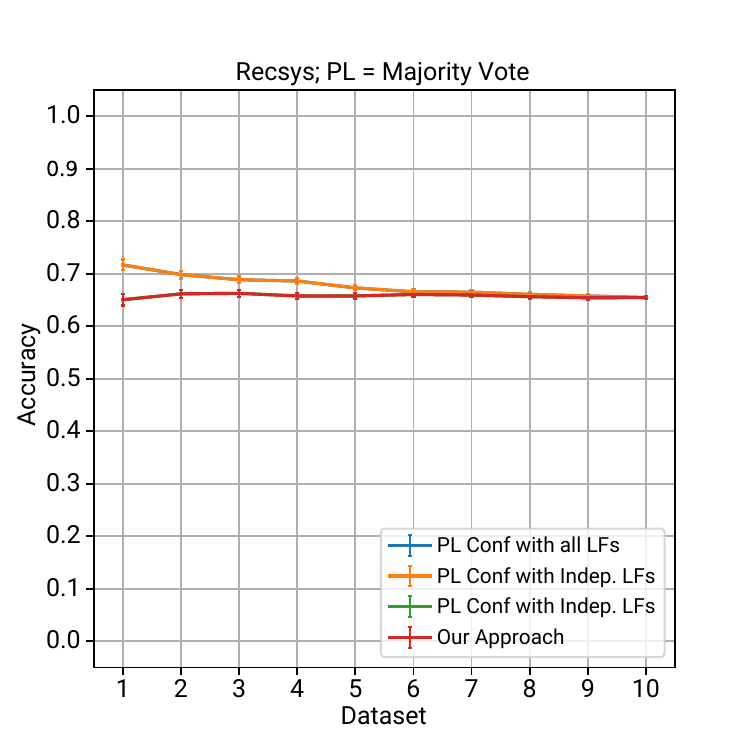}
    }
    \hfill
    \subfigure{
        \includegraphics[width=0.22\linewidth]{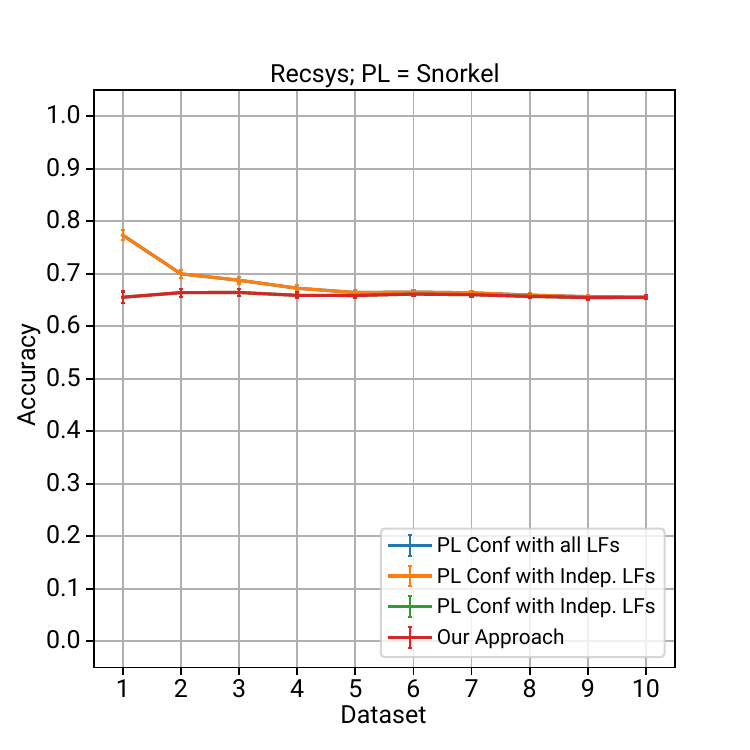}
    }
    \subfigure{
        \includegraphics[width=0.22\linewidth]{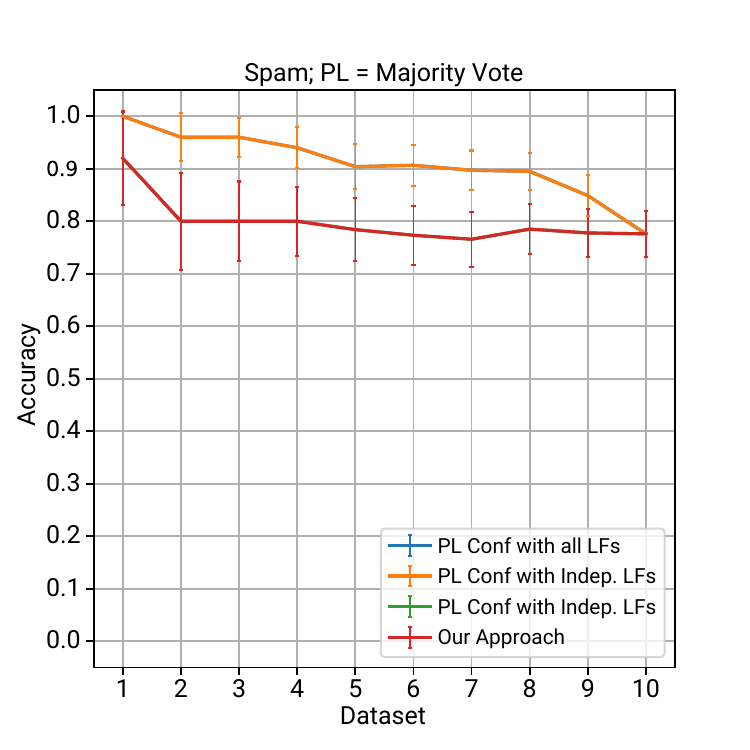}
    }
    \subfigure{
        \includegraphics[width=0.22\linewidth]{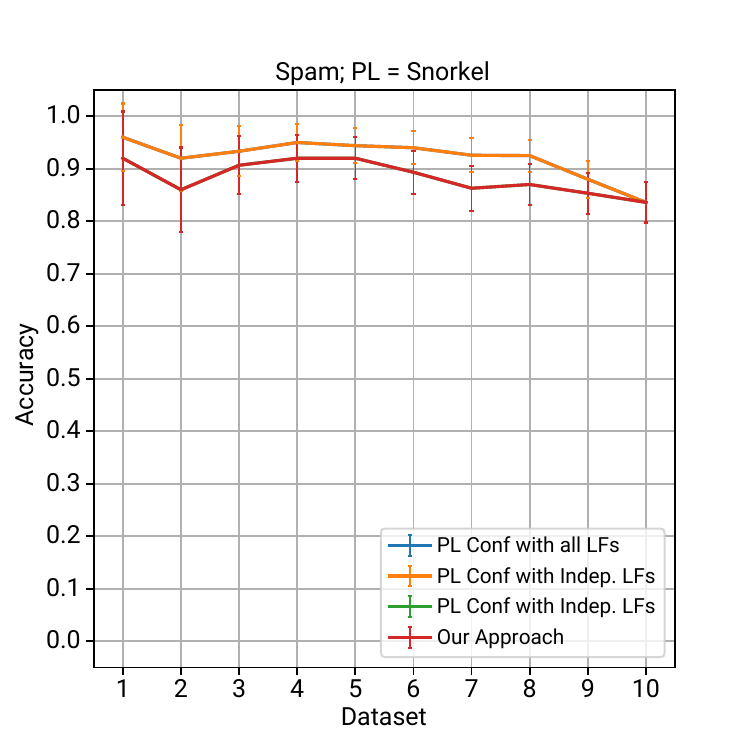}
    }
    \caption{Accuracy of datasets produced by our approach and comparative approaches. Recsys and Spam do not have dependent labeling functions hence the lines for PL with all LFs and PL with Indep. LFs, and CI LB with all LFs and our approach overlap.}
    \label{fig:topp_vs_acc}
\end{figure*}

Table~\ref{tab:apdx_rankcors} shows the full results, \ie the Spearman's Rank Correlation coefficients and p-values of the adversarially ordered natural datasets produced by our approach and comparative approaches across our nine case studies. Figure~\ref{fig:topp_vs_acc} shows the accuracy of the adversarially ordered natural datasets produced by our approach and comparative approaches across our nine case studies. For each case study, we curate ten datasets (\ie $N=10$) and then plot their accuracy scores. We also quantify the uncertainty in these accuracies by placing a 90\% binomial confidence interval around them. For each dataset $D_i = (X_i, Y_i, \Yh_i)$ where $i \in \{1,\ldots,N\}$, the width of the confidence interval is computed as, $$1.64 \cdot \sqrt{\frac{\acc(Y_i, \Yh_i) \cdot \left(1 - \acc(Y_i, \Yh_i)\right)}{\left| Y_i \right|}}$$ where $$\acc(Y, \Yh) = \frac{1}{|Y|} \sum_{i=1}^{|Y|} \ind{y_i = \yh_i}$$ and $|\cdot|$ denotes the cardinality of the given set. 

\end{document}